\documentclass[10pt,a4paper]{article}
\usepackage[margin=23mm]{geometry}
\usepackage[T1]{fontenc}
\usepackage[utf8]{inputenc}
\usepackage{lmodern,microtype}
\usepackage{amsmath,amssymb,bm}
\usepackage{graphicx,xcolor,booktabs,array,tabularx}
\usepackage{caption,placeins,enumitem}
\usepackage{tikz,pgfplots}
\pgfplotsset{compat=1.18}
\usepackage[numbers,sort&compress]{natbib}
\usepackage{xurl,hyperref}
\definecolor{inkblue}{HTML}{244C6A}
\definecolor{curveblue}{HTML}{247A91}
\definecolor{curveorange}{HTML}{D78B36}
\definecolor{gainblue}{HTML}{245E91}
\definecolor{lossred}{HTML}{A94A46}
\newcommand{\gain}[1]{\textcolor{gainblue}{$\uparrow\,#1$}}
\newcommand{\drop}[1]{\textcolor{lossred}{$\downarrow\,#1$}}
\hypersetup{colorlinks=true,linkcolor=inkblue,citecolor=inkblue,urlcolor=inkblue,
  pdftitle={Causal World Models under Physical Parameter Shifts},pdfauthor={Chu Lin and Yi Sun}}
\setlist[itemize]{leftmargin=1.6em,itemsep=0.15em,topsep=0.25em}
\newcolumntype{L}[1]{>{\raggedright\arraybackslash}p{#1}}
\newcommand{\Rbar}{\overline{R^2}}
\newcommand{\clip}{\operatorname{clip}}

\newcommand{\softplus}{\operatorname{softplus}}
\newcommand{\mse}{\operatorname{MSE}}
\newcommand{\E}{\mathbb{E}}
\begin{document}

\begin{center}
{\LARGE\bfseries Causal World Models under\\[.15em]Physical Parameter Shifts\par}
\vspace{.6em}
Chu Lin$^{1}$ and Yi Sun$^{2,*}$\\[.3em]
{\small $^{1}$Department of Computer Science, $^{2}$Department of Electrical Engineering,\\
The City College of New York, New York, NY 10031, USA}\\[.3em]
{\small $^{*}$Corresponding author: \href{mailto:ysun@ccny.cuny.edu}{ysun@ccny.cuny.edu}}
\end{center}

\begin{abstract}
World models predict how embodied systems evolve under actions, but their accuracy can degrade when physical conditions move outside the training distribution. We study two components of causal modeling: which inputs a predictor may use, and how its mechanism computes their effects. In a synthetic benchmark with shortcut features, a frozen dependency mask derived from paired interventions reduces mean out-of-distribution (OOD) RMSE from 1.769 for a fully connected predictor to 0.697, close to the generator-oracle mask (0.694). For robot link-mass extrapolation in CausalWorld, with disjoint training and test mass ranges, we develop finger-local world models in which link masses enter inertia and dynamics-bias terms through explicit affine coefficients, actions act through either a learned torque network or the known PD controller, and learned residuals capture contact-related effects. Compared with a separately trained control whose mass inputs are fixed at training means, the known-PD model turns out-of-range masses into a consistent predictive gain: median time-averaged joint $R^2$ over 150-step rollouts rises from 0.609 to 0.727, with improvement in all ten task--seed runs (exact two-sided sign test, $p\approx0.002$). In a shared 100-step forecast, the learned-torque model is competitive with the strongest recurrent robotic world model (RWM) configuration (median joint $R^2$ of 0.755--0.768 versus 0.752), whereas supplying out-of-range masses sharply degrades the MLP and recurrent predictors. These results indicate that how a physical parameter enters the dynamics mechanism determines whether its out-of-range value helps or harms prediction.
\end{abstract}


\section{Introduction}
\label{sec:intro}
World models are used to predict the evolution of embodied systems as they execute actions~\cite{li2025rwm}. Their prediction accuracy may degrade when physical conditions deviate from the training environment. Recent theory connects generalization across multi-step goals to learned predictive world models~\cite{richens2025}. Embodied applications require these models to remain useful under changing conditions, such as when a robot handles a different payload or operates with different physical properties. Pearl's structural causal model (SCM) framework separates structural dependencies from the functions governing those relationships~\cite{pearl2009}. This separation motivates two questions for a world model: which inputs should a predictor use, and how should its dynamics mechanism compute their effects?

Causal dynamics models, such as CDL~\cite{wang2022} and recent structure-distribution approaches~\cite{murillo2025}, learn or estimate dependencies between current state, action, and next state, and use them to restrict irrelevant inputs to the dynamics predictor. Such constraints remove direct prediction paths from excluded shortcut inputs. However, if the OOD shift involves a true causal factor, such as a heavier robot link, the relevant dependency must remain. The computation performed after reading the shifted value is then determined by the mechanism. Input selection alone does not specify how this function should extrapolate. Recent context-aware mechanical models and structured physical world models further motivate examining how parameter information enters the dynamics mechanism~\cite{schulze2025context,wang2026}.

\begin{figure}[!htbp]
\centering
\includegraphics[width=.96\linewidth]{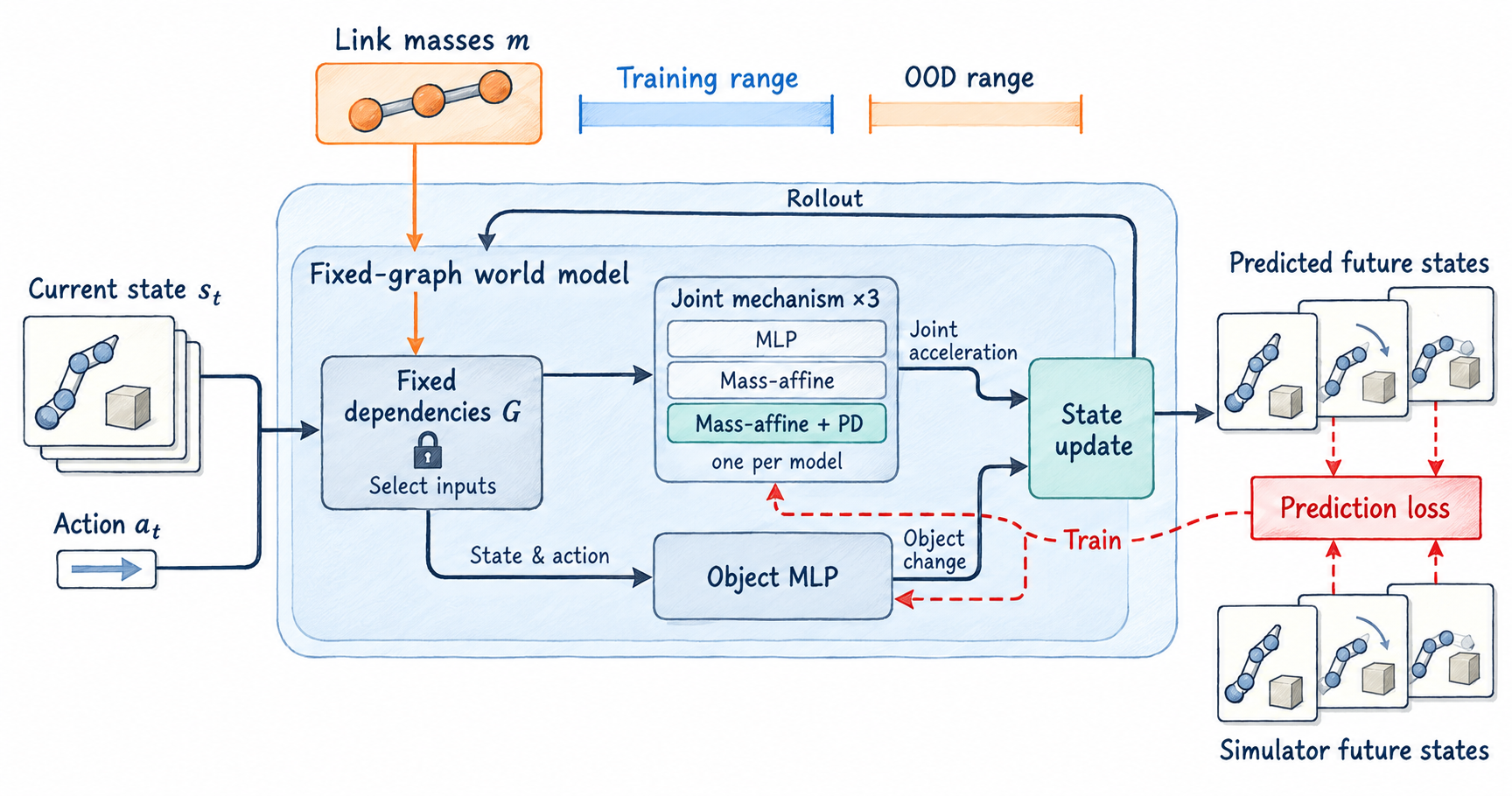}
\caption{Overview of the robot world model. Three finger mechanisms predict joint accelerations under fixed input dependencies; one mechanism family is used per model. A separate object MLP predicts object-state changes. Joint integration and object updates produce future states, which are fed back for rollout prediction. Training uses simulator trajectories; evaluation uses link masses outside the training range.}
\label{fig:overview}
\end{figure}

We develop a robot world-model configuration that assigns physical parameters and control inputs distinct roles: link masses enter inertia and dynamics bias through explicit affine coefficients, actions produce torque through the known PD controller, and a neural residual models contact-related effects without directly receiving robot masses. The dependencies are fixed to each finger's local state and action, with shared object context. We compare this configuration with an MLP and mass-affine dynamics with learned torque under the same input routing. Each model is paired with a separately trained mass-information control: mass inputs are removed from the MLP or fixed at their training means in the structured models. A complementary synthetic experiment uses an intervention-derived mask to evaluate shortcut dependence. Figure~\ref{fig:overview} summarizes the robot architecture and prediction procedure.

The experiments connect dependency selection to physical parameter extrapolation. Frozen intervention-derived masks reduce mean synthetic OOD RMSE by 60.6\%. In CausalWorld, supplying out-of-range link masses to an unstructured MLP or to a recurrent robotic world model sharply degrades prediction, whereas the known-PD mass-affine model improves its median time-averaged joint $R^2$ over 150 steps from 0.609 for its fixed-mass-input control to 0.727. The median within-run benefit of mass information is 0.121 under the heavy-mass shift, compared with 0.009 within the training mass range. In a shared 100-step benchmark, the learned-torque model is competitive with the strongest recurrent baseline (median joint $R^2$ of 0.755 with actual masses and 0.768 with fixed mass inputs, versus 0.752), and the known-PD model retains a positive mass benefit in every run.

Our contributions are summarized as follows:
\begin{itemize}
\item \textbf{A mass-informed robot world-model design.} We develop finger-local mass-affine dynamics with learned-torque and known-PD variants. Masses enter inertia and bias through explicit coefficients, while torque and residual networks use mass-excluded inputs. Joint integration and a separate object-state predictor generate future trajectories.
\item \textbf{Competitive OOD prediction in a shared CausalWorld benchmark.} Over a common 100-step forecast, the learned-torque model reaches median joint $R^2$ of 0.755 with actual masses and 0.768 with fixed mass inputs, compared with 0.752 for the strongest tested RWM configuration, with slightly lower joint NRMSE and object-position RMSE and without using observation history. The accuracy margins are small; the larger difference lies in how the models respond to out-of-range mass inputs, which sharply degrade the MLP and recurrent predictors.
\item \textbf{Consistent predictive benefit from out-of-range masses.} Separately trained mass-information controls show that the known-PD design improves median 150-step mean joint $R^2$ from 0.609 to 0.727. Actual masses improve prediction in all ten task--seed runs in both the 150-step ablation and the 100-step benchmark; the latter yields a median paired mass gain of 0.071 (exact two-sided sign test, $p\approx0.002$ in each case).
\end{itemize}

\section{Problem Statement}
\label{sec:problem}
Given state $s_t$, action $a_t$, robot link masses $m$, and other physical context $c$, we predict
\begin{equation}
\hat s_{t+1}=F_{\theta,G}(s_t,a_t,m,c),
\label{eq:worldmodel}
\end{equation}
where $G$ specifies permitted input dependencies and $\theta$ parameterizes the mechanisms. The robot mass study assumes that the current link masses and the simulator's low-level control law are available. The networks learn dynamics coefficients from state-prediction supervision, using the current mass values to specify the physical condition. A rollout starts from an observed state and repeatedly applies $F_{\theta,G}$ using predicted states and the recorded action sequence.

We distinguish two sources of distribution shift. In a \emph{shortcut shift}, an input correlated with an output during training changes its relationship to that output; excluding an irrelevant input removes that prediction path. In \emph{physical parameter extrapolation}, a relevant parameter takes values outside the training support. The model must retain its dependence on that parameter and compute its effect at unseen values. The synthetic study evaluates fixed masks in a benchmark containing shortcut shifts. The robot study isolates a link-mass shift under a shared dependency structure.

The robot evaluation asks both whether predictions are accurate and whether actual mass information improves them. For mechanism family $r$ and task--seed run $b$, we define the predictive benefit of mass as
\begin{equation}
\Delta_{r,b}=\Rbar_{r,b}^{\mathrm{actual}}-\Rbar_{r,b}^{\mathrm{control}},
\label{eq:benefit}
\end{equation}
where $\Rbar$ is the mean joint $R^2$ across the evaluated rollout steps. The controls and score aggregation are specified below. This separates absolute prediction accuracy from the additional value of providing the physical parameter.

\section{Related Work}
\label{sec:related}
\textbf{Causal dynamics and dependency constraints.}
SCMs separate dependencies from the mechanisms generating variables~\cite{pearl2009}. Theoretical studies connect robustness under specified distribution shifts to causal world-model knowledge, including settings in which actions affect the environment~\cite{richens2024,ceriscioli2025}. CDL learns causal input dependencies for dynamics prediction and task-independent state abstraction~\cite{wang2022}; FCDL refines the structure through quantized state--action subgroups~\cite{hwang2024}. More recently, CADY estimates a distribution over causal structures and samples masks for a probabilistic dynamics model, evaluating robustness to corrupted inputs and changes in robot dynamics~\cite{murillo2025}. CoDA uses local causal structure for counterfactual data augmentation~\cite{pitis2020}, while BECAUSE learns causal representations to address confounding in offline model-based reinforcement learning~\cite{lin2024}. We fix the dependency structure to study the functional response to a supplied physical parameter outside its training range.

\textbf{Composable and masked world models.}
WM3C uses language-guided compositional causal components to model unseen environments~\cite{wang2025wm3c}. Causal-JEPA uses object-level latent masking to encourage interaction-dependent prediction and evaluates visual reasoning and predictive control~\cite{nam2026}. These approaches organize latent information and interactions. Our study uses recorded physical states and known link masses, focusing on how a dynamics mechanism turns an out-of-range parameter into a useful change in its predicted trajectory.

\textbf{Structured mechanics and physical parameterization.}
DeLaN embeds Lagrangian mechanics into learned dynamics~\cite{lutter2019,lutter2023}; classical inertial identification uses the linear dependence of inverse dynamics on inertial parameters~\cite{atkeson1986}. Structured Mechanical Models and APHYNITY combine known physical structure with learned components~\cite{gupta2020,yin2021}. Recent extensions are particularly relevant: Context-Aware DeLaN combines recurrent online identification with contextual mechanical models and residual dynamics for a robot arm under varying loads~\cite{schulze2025context}; Floating-Base DeLaN imposes branch-induced sparsity and physical consistency on learned inertia matrices~\cite{schulze2025floating}. Both directly build on DeLaN. Our configuration makes measured masses explicit affine coefficients of inertia and bias, routes actions through a known controller, and assesses parameter utility using separately trained mass-information controls.

\textbf{Physical world models and extrapolation.}
Architectural functional form influences neural extrapolation~\cite{xu2021}. Wang et al.\ study structured physical world models with learned energy, dissipation, known drive channels, and separable intervention-parameter maps in electromagnetic and fluid systems~\cite{wang2026}. This combines imposed physical structure with learned system-specific functions. We study contact-rich robot prediction under link-mass changes, using a mass-affine joint model with a separately learned state- and velocity-dependent bias and an object-state predictor. The target is the predictive benefit of known masses over finite robot rollouts.

\textbf{History-based adaptation and robotic prediction.}
UP-OSI estimates physical parameters from state--action history, while CaDM conditions dynamics on a learned context; RMA uses an adaptation module for policy control~\cite{yu2017,lee2020,kumar2021}. Context-Aware DeLaN further couples online identification to structured mechanics~\cite{schulze2025context}. RWM uses recurrent history and autoregressive training for long-horizon robotic prediction~\cite{li2025rwm}. These methods motivate separating inference of unknown context from use of a supplied parameter: in our mass experiments the actual link masses and low-level control law are available, and the question is how the dynamics representation uses them.

\textbf{Evaluation and baseline methods.}
PETS combines probabilistic ensemble dynamics, trajectory sampling, and model-predictive control~\cite{chua2018}. CausalWorld provides configurable robot and object properties for embodied transfer studies~\cite{ahmed2021}. We use CausalWorld for both environment-transfer prediction with a PETS-style dynamics baseline and the link-mass mechanism study, with separate metrics for the two protocols. Appendix~\ref{app:origins} identifies the sources of established components and our design choices.

\section{Methodology}
\label{sec:methods}
\subsection{Fixed dependencies and model decomposition}
The robot world model contains three separately parameterized finger mechanisms and one object-state MLP. Each finger uses its own three joint positions, three velocities, three actions, four mass entries, and shared object state and context. Its mechanism predicts three joint accelerations, which are integrated to update that finger's joint state. The object MLP reads all joint positions, velocities, and actions together with the object state and context, and predicts a 13-dimensional state increment. It receives robot-mass effects through the evolving state. This input routing is fixed across the three mechanism families; each model learns its own network weights.

For the synthetic dependency experiment, the mask is estimated from paired interventions. For each intervened input $i$, we average the absolute output change and normalize each output column:
\begin{equation}
E_{ij}=\frac{1}{N_i}\sum_{n:\,I_n=i}|y^{\mathrm{cf}}_{n,j}-y^{\mathrm{base}}_{n,j}|,
\quad S_{ij}=\frac{E_{ij}}{\max_kE_{kj}+10^{-9}},
\quad A_{ij}=\mathbf{1}\{S_{ij}\geq10^{-3}\max_{k,l}S_{kl}\}.
\label{eq:mask}
\end{equation}
Scores use the training intervention pool; the mask is frozen before fitting the predictor. A shared scalar encoder, edge-specific linear maps, and output heads give
\begin{equation}
\hat y_j=g_{\theta,j}\!\left(\sum_i A_{ij}W_{ij}\phi_\theta(x_i)\right).
\end{equation}
The robot structure is specified by finger-local routing, whereas the synthetic mask is intervention-derived.

\subsection{Mass-affine dynamics mechanisms}
\label{sec:massaffine}
The MLP baseline maps the permitted inputs to joint acceleration. The structured alternatives separate the known mass values from learned state-dependent coefficients. For one finger, let $\mu_r=m_r/(0.03\,\mathrm{kg})$. We form
\begin{align}
\mathcal M(q,m)&=L_0(q)L_0(q)^\top+\sum_{r=1}^{4}\mu_r L_r(q)L_r(q)^\top+10^{-4}I,\label{eq:inertia}\\
h_\theta(q,\dot q,m)&=b_0(q,\dot q)+\sum_{r=1}^{4}\mu_r b_r(q,\dot q).\label{eq:bias}
\end{align}
Each $L_r$ is lower triangular with diagonal entries $\softplus(v)+10^{-3}$, giving positive-definite $\mathcal M$ for nonnegative masses. The bias coefficients are learned separately from the inertia coefficients. $h_\theta$ models position- and velocity-dependent dynamics bias. All coefficient networks depend on state, while mass enters through the explicit coefficients $\mu_r$.

The learned-torque configuration computes
\begin{equation}
\ddot q=D_\alpha\mathcal M^{-1}
\bigl[\tau_\theta(q,\dot q,a)+c_\theta(q,\dot q,z)-h_\theta\bigr],
\label{eq:learnedtorque}
\end{equation}
where $z$ contains object state and context, and $D_\alpha$ is a learned positive diagonal acceleration scale initialized from training acceleration statistics. The torque network and contact-related residual $c_\theta$ do not directly take robot mass as input. The residual is learned end to end from state-prediction error. The terms $\mathcal M$ and $h_\theta$ are affine in mass; solving the dynamics introduces the corresponding nonlinear dependence of acceleration on mass.

\subsection{Known PD control and state updates}
\label{sec:controller}
The known-PD configuration maps normalized action $a$ to target joint angles $q^*$ using the simulator's joint limits, and computes
\begin{equation}
\tau_{\mathrm{PD}}=\clip_{\tau_{\max}}\!\left[
\clip_{\tau_{\max}}\bigl(K_p(q^*-q)-K_d\dot q\bigr)-K_{d,s}\dot q\right].
\label{eq:pd}
\end{equation}
Clipping is elementwise. The controller uses $K_p=10$, $K_d=(0.1,0.3,0.001)$, $K_{d,s}=(0.08,0.08,0.04)$ per finger, and $\tau_{\max}=0.36$ N\,m. This reproduces the simulator's PD torque, safety damping, and torque limits~\cite{causalworldcode}. The rule uses action and joint state; mass affects its rollout response through state feedback. The mass-affine dynamics in physical units are
\begin{equation}
\ddot q=(S\mathcal M S)^{-1}
\bigl[\tau_{\mathrm{PD}}+\tau_{\max}c_\theta(q,\dot q,z)-\tau_{\max}h_\theta\bigr],
\label{eq:pddynamics}
\end{equation}
where the fixed diagonal scale $S$ is the square root of the per-joint ratio of training torque standard deviation to training acceleration standard deviation. Actions enter this mechanism through the known controller; networks learn the inertia, dynamics bias, and contact-related residual. This configuration specifies both the torque source and the physical scaling of the learned terms.

For all mechanisms, accelerations are softly limited, followed by semi-implicit integration:
\begin{equation}
\tilde{\ddot q}=a_{\max}\tanh(\ddot q/a_{\max}),\quad
\hat{\dot q}_{t+1}=v_{\max}\tanh\!\left(\frac{\dot q_t+\Delta t\,\tilde{\ddot q}}{v_{\max}}\right),\quad
\hat q_{t+1}=q_t+\Delta t\,\hat{\dot q}_{t+1}.
\label{eq:update}
\end{equation}
Limits and arithmetic are coordinatewise. The object MLP predicts a normalized increment that is transformed back to state units and added to the current object state. Object quaternions are normalized during rollouts. Joint and object predictions form the next state in Eq.~\eqref{eq:worldmodel}.

\subsection{Training and mass-information controls}
Models are first trained on one-step transitions and then fine-tuned on free-running trajectories. The multi-step objective combines teacher-forced one-step error with rollout error:
\begin{equation}
\mathcal L=\E\,\mse\!\left(\frac{\hat s_{t+1}^{\mathrm{TF}}-s_{t+1}}{\sigma_{\Delta s}}\right)
+\frac{\lambda}{H}\sum_{k=1}^{H}\E\,\mse\!\left(\frac{\hat s_{t+k}^{\mathrm{roll}}-s_{t+k}}{\sigma_s}\right),
\label{eq:loss}
\end{equation}
where $\mse$ averages squared components, $\lambda=1$, and normalization statistics are fitted to training data. Each rollout begins with an observed state and then uses its own predictions. Model selection uses validation trajectories from the training mass domain.

We use three complementary mass-information controls. First, a separately trained MLP removes robot-mass inputs. Second, each structured model has a separately trained counterpart in which mass inputs are overwritten by their training means during both training and evaluation. These counterparts use the same varying-mass simulator trajectories as the corresponding actual-mass models. Third, an inference-time diagnostic replaces the trained MLP's test mass inputs with training means while retaining its weights. This last comparison measures the effect of the shifted input on an already fitted predictor.

\section{Simulation Results}
\label{sec:results}
\subsection{Experimental setup and metrics}
The synthetic benchmark has two interacting objects, 32 inputs, six next-state outputs, and 17 distractor inputs. Its OOD split changes physical inputs and shortcut relationships. We compare dense, random, intervention-derived, and generator-oracle masks using the same predictor architecture, data, and loss. Random masks match the oracle's edge count. Inputs and outputs are standardized using training statistics, and RMSE is computed in those standardized output coordinates over ten seeds.

The robot experiment uses CausalWorld pushing and picking tasks~\cite{ahmed2021}. Nine movable-link masses vary, while three base masses remain fixed, giving 12 mass inputs. Each varying mass is sampled within its simulator space A bounds $(\ell_r,h_r)$ as
\begin{equation}
m_r=\ell_r+u_r(h_r-\ell_r),\qquad
u_r\sim U(0,0.4)\ \text{for training},\quad
u_r\sim U(0.6,1)\ \text{for OOD testing}.
\label{eq:domains}
\end{equation}
The mean test-to-training mass ratio is approximately 1.85. Both domains are inside platform space A, with a gap between the light and heavy training/test subranges. Per task--seed run, 1,200 light-mass episodes are split into 864 training, 96 validation, and 240 in-distribution (ID) test episodes; 300 separate heavy-mass episodes form the OOD test. Each episode has 150 steps at 250 Hz, and action targets change every 50 steps.

The 31-dimensional state contains nine joint positions, nine velocities, and the object's position, quaternion, linear velocity, and angular velocity. Other context consists of object mass, three size coordinates, and object, floor, and stage friction. Network inputs are standardized using training statistics; the PD rule uses joint positions and velocities in physical units. Finger and coefficient networks have two SiLU hidden layers of width 128; the object MLP has three hidden layers of width 256. Acceleration limits use ten times the training 0.999 quantile of absolute velocity increments divided by $\Delta t$; velocity limits use three times the maximum absolute training velocity. Table~\ref{tab:training} gives the robot training settings.

\begin{table}[htbp]
\centering\small
\caption{Robot training settings. Actual-mass and fixed-mass-input PD models use the same candidate grid and select their settings separately on light-mass validation data.}
\label{tab:training}
\begin{tabularx}{\linewidth}{L{.22\linewidth}XXX}
\toprule
Setting & One-step training & MLP / learned torque rollout & Known-PD rollout\\
\midrule
Optimizer / batch & Adam / 256 & Adam / 128 & Adam / 128\\
Learning rate & $10^{-3}$ & $3\times10^{-4}$ & $\{10^{-4},3\times10^{-5}\}$\\
Horizon & 1 & 50 & 10, then 50\\
Training budget & 80 epochs & 1,500 updates & 1,500 updates per stage\\
Validation interval / patience & Every epoch / 10 & Every 100 updates / 5 & Every 25 updates / 8\\
\bottomrule
\end{tabularx}
\end{table}

One-step learning rates are halved after three validation plateaus; the gradient-norm limit is 1.0. Rollout validation uses 512 windows. Within the PD family, actual-mass and fixed-mass-input counterparts begin one-step training from identical weights and matched minibatch order; each final model is selected by its own validation loss. The configurations in Table~\ref{tab:training} define the mechanism comparisons.

For each rollout step $k$, joint coordinates are scaled by their training standard deviations. With $z_{n,k,j}$ denoting scaled coordinate $j$ in test episode $n$ and $\bar z_{k,j}$ its mean over test episodes, we compute
\begin{equation}
R_k^2=1-\frac{\sum_{n,j}(\hat z_{n,k,j}-z_{n,k,j})^2}{\sum_{n,j}(z_{n,k,j}-\bar z_{k,j})^2},\qquad
\Rbar=\frac{1}{H}\sum_{k=1}^{H}R_k^2.
\label{eq:metric}
\end{equation}
At each step, $R_k^2=1$ denotes exact joint-state prediction, $R_k^2=0$ matches the per-step test-mean baseline, and a negative value indicates greater squared error than that baseline. The 18 coordinates include joint positions and velocities. The mass-information ablation uses $H=150$ from the episode's initial state; the RWM comparison uses a common $H=100$ forecast after an observed history. We also report joint normalized RMSE (NRMSE) and object-position RMSE:
\begin{equation}
\operatorname{NRMSE}_{\mathrm{joint}}
=\sqrt{\frac{1}{18EH}\sum_{n=1}^{E}\sum_{k=1}^{H}\sum_{j=1}^{18}
\left(\frac{\hat s_{n,k,j}-s_{n,k,j}}{\sigma_j}\right)^2},\qquad
\operatorname{RMSE}_{\mathrm{obj}}
=\sqrt{\frac{1}{3EH}\sum_{n=1}^{E}\sum_{k=1}^{H}
\|\hat p_{n,k}-p_{n,k}\|_2^2},
\label{eq:errors}
\end{equation}
where $E$ is the number of test episodes, $\sigma_j$ is a training-set joint-state standard deviation, and $p$ is the object's three-dimensional position. Joint NRMSE expresses joint-state error in training-standard-deviation units, making position and velocity errors comparable on a dimensionless scale. Object RMSE measures the root-mean-square error over the three position coordinates and is converted from meters to centimeters in the comparison table. Higher $\Rbar$ and lower errors indicate better predictions.

We report medians of run-level metrics over two tasks and five seeds. Mass benefit is computed within each run using Eq.~\eqref{eq:benefit}, then summarized by $\Delta_{\mathrm{mass}}=\operatorname{median}_b\Delta_{r,b}$ and the number of positive differences. A gain against a named baseline uses the same paired aggregation. A median paired gain can differ from the difference of the two model-score medians. These gains are changes in $R^2$ units, not percentage improvements. Counts are descriptive: task runs within a seed share mass and action sampling streams. Where we summarize the direction of paired differences, we also report an exact two-sided sign test over the ten runs (10/10 positive differences give $p\approx0.002$; 8/10 give $p\approx0.11$); because runs within a seed are not fully independent, these $p$-values are indicative only. Absolute $R^2$ values are normalized by the variance of each test split, so comparisons between the ID and OOD splits rely on paired within-split differences rather than on absolute scores. The synthetic standardized RMSE and the environment-transfer skill score use their own targets and definitions, specified with their respective results.

\subsection{Prediction with fixed dependency masks}
Table~\ref{tab:synthetic} shows that the intervention-derived mask reduces mean OOD RMSE from 1.769 to 0.697, a 60.6\% reduction, with improvement in all ten seeds. The oracle-mask error is 0.694. The random sparse mask has an error of 1.604, indicating that the selected dependency pattern matters beyond applying a generic sparse mask. These comparisons establish the value of the frozen dependency constraint for prediction in this synthetic setting.

\begin{table}[htbp]
\centering\small
\caption{Synthetic OOD prediction. Values are ten-seed means in training-standardized output coordinates. Lower is better.}
\label{tab:synthetic}
\begin{tabular}{lc}
\toprule
Input mask & OOD RMSE\\
\midrule
Dense & 1.769\\
Random sparse & 1.604\\
Intervention-derived, frozen & \textbf{0.697}\\
Generator oracle reference & 0.694\\
\bottomrule
\end{tabular}
\end{table}

The predictor uses a 16-dimensional scalar encoding, 16-dimensional edge messages, encoder hidden width 64, and output-head hidden width 128. All mask variants use squared prediction error and the same effect-normalized paired-response loss. Adam uses learning rate $3\times10^{-3}$, batch size 2,048, weight decay $10^{-6}$, and at most 300 epochs with validation patience 40. The paired-loss weight increases from 0.25 to 1 after 20 epochs. Validation combines observational prediction and paired-response scores. This holds the learning procedure fixed while comparing the available input dependencies.

\subsection{Mass-information ablation across mechanism configurations}
\label{sec:massresults}
Table~\ref{tab:mass} ablates mass information within each mechanism family under the heavy-mass shift. The known-PD model uses the actual mass to improve its median score from 0.609 to 0.727. The controls retain the same varying-mass trajectories and remove or fix the mass supplied to the predictor, so the comparison measures the value of that input within each family. The MLP's separately trained no-mass control obtains 0.721, compared with $\Rbar=-0.803$ with actual mass. Replacing test masses with training means in the fitted actual-mass MLP gives 0.717. This diagnostic shows why input availability and useful parameter dependence are evaluated separately.

\begin{table}[htbp]
\centering\small
\caption{Mass-information ablation under link-mass extrapolation. Values are medians of 150-step mean joint $R^2$ over two tasks and five seeds. Within each row, the actual-mass model is compared with a separately trained control of the same mechanism family. Higher is better.}
\label{tab:mass}
\begin{tabularx}{\linewidth}{XccL{.30\linewidth}}
\toprule
Mechanism & Actual mass & Ablated mass & Ablation definition\\
\midrule
MLP & $-0.803$ & 0.721 & Mass inputs removed\\
Mass-affine, learned torque & 0.734 & 0.733 & Mass inputs fixed at training means\\
Mass-affine, known PD & 0.727 & 0.609 & Mass inputs fixed at training means\\
\bottomrule
\end{tabularx}
\vspace{.25em}
\begin{minipage}{\linewidth}\footnotesize
All controls use the same varying-mass simulator data as their actual-mass counterparts; only access to mass information is changed. The structured controls use training-mean masses in training and testing; the MLP control removes mass inputs. Validation selects settings within each family as specified in Table~\ref{tab:training}. The inference-only MLP substitution score, 0.717, is a separate diagnostic.
\end{minipage}
\end{table}

The mass-affine known-PD model obtains a median $\Rbar$ of 0.727, compared with 0.609 for its fixed-mass-input counterpart. Actual mass improves $\Rbar$ in all ten task--seed runs, with a mean within-run gain of 0.120 (median 0.121; exact two-sided sign test, $p\approx0.002$). The learned-torque model achieves medians of 0.734 and 0.733 for actual and fixed mass, respectively, with positive within-run mass benefit in three of ten runs. The known-PD configuration consistently converts supplied out-of-range mass into predictive benefit under this protocol. Its fixed-mass-input control is, however, markedly weaker than the learned-torque control (0.609 versus 0.733). One plausible explanation is that, with torque fixed by the known controller, the inertia and bias terms are the only channels through which the model can represent heavier links, so the PD mechanism relies on mass information more than the learned-torque model does. The mass benefit should therefore be read together with the absolute scores. Comparisons between mechanism families concern complete configurations, including their scaling and training settings; the matched controls quantify the benefit of mass information within each family.

\subsection{Mass benefit across domains and rollout horizons}
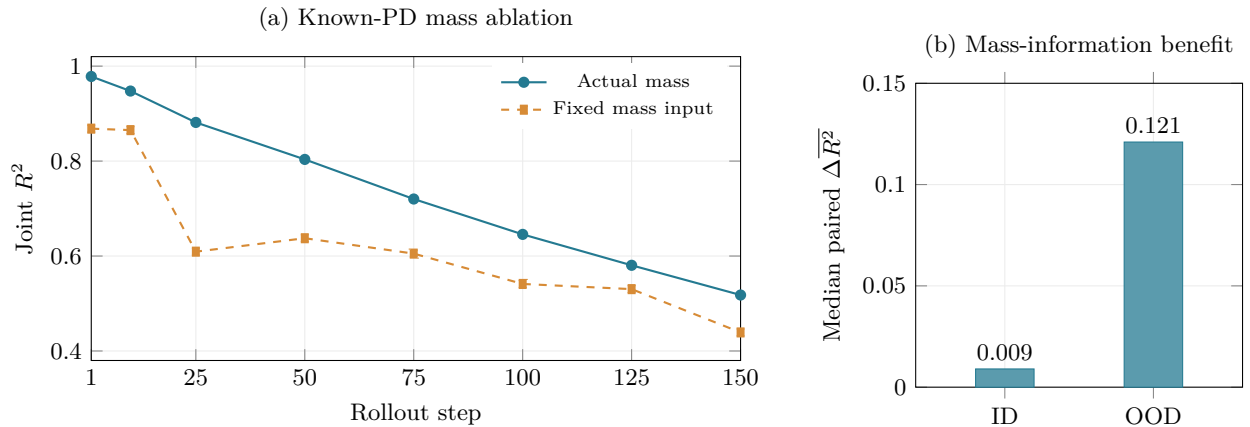
\begin{figure}[htbp]
\centering
\begin{tikzpicture}
\begin{axis}[
 width=.62\linewidth,height=5.6cm,
 xlabel={Rollout step},ylabel={Joint $R^2$},
 xmin=1,xmax=150,ymin=.38,ymax=1.02,
 xtick={1,25,50,75,100,125,150},ytick={.4,.6,.8,1},
 tick label style={font=\small},label style={font=\small},
 grid=major,grid style={gray!15},
 legend style={at={(.98,.98)},anchor=north east,font=\scriptsize,draw=none,fill=white},
 title={(a) Known-PD mass ablation},title style={font=\small},
 ]
 \addplot[curveblue,thick,mark=*,mark size=1.7pt] coordinates
 {(1,.9782)(10,.9475)(25,.8814)(50,.8035)(75,.7201)(100,.6456)(125,.5805)(150,.5179)};
 \addlegendentry{Actual mass}
 \addplot[curveorange,thick,dashed,mark=square*,mark size=1.5pt] coordinates
 {(1,.8683)(10,.8652)(25,.6092)(50,.6375)(75,.6051)(100,.5412)(125,.5303)(150,.4391)};
 \addlegendentry{Fixed mass input}
\end{axis}
\end{tikzpicture}\hfill
\begin{tikzpicture}
\begin{axis}[
 width=.36\linewidth,height=5.6cm,
 ybar,bar width=22pt,ymin=0,ymax=.15,
 symbolic x coords={ID,OOD},xtick=data,ytick={0,.05,.10,.15},
 ylabel={Median paired $\Delta\Rbar$},
 tick label style={font=\small},label style={font=\small},
 grid=major,grid style={gray!15},enlarge x limits=.6,
 title={(b) Mass-information benefit},title style={font=\small},
 nodes near coords,nodes near coords style={font=\small,/pgf/number format/.cd,fixed,precision=3},
 yticklabel style={/pgf/number format/.cd,fixed,precision=2},
 ]
 \addplot[fill=curveblue!75,draw=curveblue] coordinates {(ID,.009)(OOD,.121)};
\end{axis}
\end{tikzpicture}
\caption{Predictive value of actual mass within the known-PD mechanism. (a) Points are medians of per-step scores across the ten task--seed runs; lines connect recorded evaluation steps. (b) Bars show medians of within-run differences in 150-step mean joint $R^2$. These paired differences are aggregated separately from the model-score medians in Table~\ref{tab:mass}.}
\label{fig:benefit}
\end{figure}

Figure~\ref{fig:benefit} shows the known-PD mass ablation over the full prediction horizon. Actual mass has a higher median score than fixed mass at every reported step. At step 150, the medians are 0.518 and 0.439, respectively. The positive benefit therefore extends to the end of the evaluated rollout.

The benefit is larger when the parameter moves outside the training range. Taking the actual-minus-control difference within each run and then its median gives 0.121 on heavy-mass episodes and 0.009 on ID episodes. The OOD benefit exceeds the ID benefit in every task--seed run. This connects the result to the intended application: when the robot's physical parameters change beyond the observed range, the structured mechanism makes the known parameter useful for trajectory prediction.

\subsection{Comparison with recurrent robotic world models}
\label{sec:rwmresults}
We compare the structured mechanisms with Robotic World Model (RWM)~\cite{li2025rwm}, a recurrent predictor trained with autoregressive state feedback. We adapt its official dynamics implementation to the same CausalWorld state, action, and context variables. RWM uses a two-layer GRU with 256 hidden units and Gaussian prediction heads. We evaluate training forecast lengths of 8 and 50 steps, each with actual-mass and no-mass inputs. This evaluates recurrent prediction alongside explicit physical parameterization; Appendix~\ref{app:rwm} gives the adaptation and training settings.

Every model predicts the same states $s_{33},\ldots,s_{132}$ from the observed state $s_{32}$ and recorded actions $a_{32},\ldots,a_{131}$. RWM additionally receives the preceding observed history. The forecast contains 100 steps at 250\,Hz (0.4\,s), with predicted states fed back throughout. All models use the same episode splits and training-derived normalization. Existing structured-model checkpoints are evaluated on this shared window, and RWM checkpoints are selected using ID validation. Table~\ref{tab:rwm} reports all evaluated configurations. Comparisons across families evaluate complete configurations with their respective history, controller knowledge, scaling, and training settings; within-family controls quantify mass-information benefit.

The learned-torque mass-affine model matches or slightly improves all three reported median prediction metrics relative to both no-mass RWM configurations, by margins of 0.003--0.016 in median $\Rbar$. With actual masses, it achieves $\Rbar=0.755$, joint NRMSE of 0.449, and object-position RMSE of 2.668\,cm, compared with 0.752, 0.457, and 2.842\,cm for RWM trained with a 50-step forecast. Its fixed-mass-input variant obtains the best absolute scores in Table~\ref{tab:rwm}: 0.768, 0.440, and 2.650\,cm. Given these small margins, we interpret the learned-torque configuration as competitive with the strongest recurrent baseline, while using no observation history; its accuracy does not depend on access to the actual masses.

The known-PD mechanism converts out-of-range mass into a positive predictive gain in all ten task--seed runs: median $\Rbar$ rises from 0.677 with fixed mass inputs to 0.746 with actual masses, and the median paired gain is 0.071. Relative to RWM without mass inputs trained with an eight-step forecast, the actual-mass known-PD model improves $\Rbar$ by a median paired 0.027, with a positive difference in eight of ten runs. Its median joint NRMSE is 0.461, compared with 0.478 for that RWM configuration. Extending RWM's training forecast to 50 steps raises its no-mass median $\Rbar$ to 0.752. The paired mass comparisons identify the distinctive result of the known-PD design: actual masses improve prediction consistently, while supplying them to the recurrent predictors or the learned-torque model does not yield a positive median mass benefit in this benchmark.

\begin{table}[!htbp]
\centering\small
\caption{CausalWorld link-mass extrapolation: a shared 100-step forecast after 32 observed history steps. Entries are medians over ten task--seed runs. $N_{\mathrm{tr}}$ denotes RWM's training forecast length. The final column is the median paired gain from actual mass relative to the control immediately above it. Bold marks the best value in each metric column.}
\label{tab:rwm}
\setlength{\tabcolsep}{4pt}
\begin{tabularx}{\linewidth}{Xlcccc}
\toprule
Model & Mass input & $\Rbar\uparrow$ & Joint & Object RMSE & $\Delta_{\mathrm{mass}}\uparrow$\\
& & & NRMSE$\downarrow$ & (cm)$\downarrow$ & \\
\midrule
Hold-state reference & None & $-0.320$ & 1.055 & 3.835 & ---\\
\midrule
RWM ($N_{\mathrm{tr}}=8$) & None & 0.718 & 0.478 & 2.913 & ---\\
& Actual & $-0.648$ & 1.182 & 7.311 & \drop{1.367}\\
\midrule
RWM ($N_{\mathrm{tr}}=50$) & None & 0.752 & 0.457 & 2.842 & ---\\
& Actual & $-0.540$ & 1.136 & 5.744 & \drop{1.293}\\
\midrule
Mass-affine, learned torque (A) & Mean & \textbf{0.768} & \textbf{0.440} & \textbf{2.650} & ---\\
& Actual & 0.755 & 0.449 & 2.668 & \drop{0.010}\\
\midrule
Mass-affine, known PD (B) & Mean & 0.677 & 0.520 & 2.830 & ---\\
& Actual & 0.746 & 0.461 & 2.816 & \gain{\mathbf{0.071}}\\
\bottomrule
\end{tabularx}
\vspace{.25em}
\begin{minipage}{\linewidth}\footnotesize
All models share test episodes, forecast origins, future actions, and scoring. RWM additionally uses the observed history; B uses the known controller. RWM ``None'' zeros the standardized robot-mass inputs; ``Mean'' fixes masses during training and evaluation. Positive mass gains occur in 0/10, 0/10, 2/10, and 10/10 runs for RWM (8), RWM (50), A, and B, respectively. Arrows show the direction and magnitude of paired mass effects, not changes between the displayed medians. The 150-step ablation in Table~\ref{tab:mass} starts from the episode's initial state.
\end{minipage}
\end{table}

\subsection{Environment-transfer comparison with a probabilistic ensemble baseline}
\label{sec:benchmark}
To test dependency constraints with an established dynamics-model family, we compare an intervention-masked predictor with a PETS-style probabilistic ensemble~\cite{chua2018} on CausalWorld pushing and picking. This complementary benchmark trains on simulator space A and tests on space B. Each task--seed run uses 8,000 source examples and 4,000 target examples, with a 20\% source holdout for ID testing and an independent source intervention pool for mask estimation. Each prediction maps an initial physical description and a nine-dimensional action to the simulator observation after holding that action for 50 environment steps. Data collection retains examples in which the object moves at least 5\,mm; action pools are constructed separately in each domain using the same movement criterion. Appendix~\ref{app:benchmark} specifies the predictor settings and score.

Table~\ref{tab:pets} reports mean target-domain prediction skill across ten seeds per task. Restricting the ensemble's inputs using the intervention-derived graph raises mean skill from 0.1407 to 0.5335. The predictor with output-specific masks obtains 0.6122. This comparison shows that useful dependency selection extends to a probabilistic ensemble, alongside the masked predictor. The ensemble uses PETS-style bootstrapped Gaussian dynamics networks, and the comparison evaluates their predictive dynamics. The score and 50-step transition protocol differ from the joint-rollout mass ablation in Table~\ref{tab:mass}.

\begin{table}[htbp]
\centering\small
\caption{External dynamics-baseline comparison on CausalWorld space A$\rightarrow$B. Values are mean prediction skill relative to the source-training mean, $\mathcal S$ (higher is better), over ten seeds per task. All rows share the same data, targets, and scoring procedure. This is a complementary environment-transfer benchmark.}
\label{tab:pets}
\begin{tabularx}{\linewidth}{Xccc}
\toprule
Predictor & Pushing & Picking & Overall\\
\midrule
PETS-style probabilistic ensemble & 0.2321 & 0.0492 & 0.1407\\
PETS-style ensemble + graph input selection & 0.6831 & 0.3839 & 0.5335\\
Intervention-masked predictor & 0.7508 & 0.4737 & 0.6122\\
\bottomrule
\end{tabularx}
\end{table}

\section{Conclusions and Future Work}
We developed a robot world-model design combining mass-affine dynamics, alternative torque mechanisms, and learned residuals under fixed finger-local dependencies. Frozen intervention-derived masks improve synthetic OOD prediction. In the shared CausalWorld benchmark, the learned-torque configuration is competitive with the strongest tested RWM configuration, with or without mass inputs. Supplying out-of-range masses sharply degrades the MLP and recurrent predictors, whereas the known-PD configuration makes actual mass information consistently useful: median 150-step mean joint $R^2$ increases from 0.609 to 0.727 relative to its fixed-mass-input control, and positive mass benefit is observed in all ten runs of both the 150-step ablation and the 100-step benchmark.

The present evaluation assumes available link masses and a known low-level controller for the PD variant, and covers forecasts of 0.4--0.6\,s. The robot dependency structure is specified by finger-local routing rather than learned, and the evaluation uses two tasks with five seeds each. Longer deployments, uncertain parameter estimates, and downstream control remain to be evaluated. Future work will examine intervention-based learning objectives and energy-consistent relationships between inertia and bias, with matched torque-model comparisons to isolate how controller knowledge affects mass use. These results show that explicit mechanism design can turn out-of-range physical parameters into useful information for embodied prediction.

\FloatBarrier

\clearpage
\appendix
\makeatletter
\renewcommand\@seccntformat[1]{\ifnum\pdfstrcmp{#1}{section}=0 Appendix\ \fi\csname the#1\endcsname\quad}
\makeatother

\section{Component Sources and Design Choices}
\label{app:origins}
The design uses established dependency selection, positive-definite matrix parameterization, affine inertial dependence, hybrid modeling, and PD control. Our configuration combines these components through finger-local routing, explicit mass entry in inertia and bias, mass-excluded torque and residual-network inputs, and the known simulator controller. The mass-information controls assess this configuration's use of the supplied physical parameter. Table~\ref{tab:origins} separates component sources from the design and evaluation choices in this study.

\begin{table}[htbp]
\centering\small
\caption{Sources of established components and this study's specific design and evaluation choices.}
\label{tab:origins}
\begin{tabularx}{\linewidth}{L{.19\linewidth}L{.29\linewidth}X}
\toprule
Component & Established foundation & Work developed in this study\\
\midrule
Input dependencies & Causal dynamics and dependency selection~\cite{wang2022,hwang2024} & Frozen intervention masks in the synthetic benchmark; shared finger-local constraints in robot models.\\
Dynamics mechanisms & Positive-definite inertia parameterization and inertial identification~\cite{lutter2019,atkeson1986,gupta2020} & Mass-affine inertia and state-dependent bias; torque and contact-related residual networks receive no direct mass input.\\
Known control & Hybrid modeling and simulator PD control~\cite{yin2021,gupta2020,ahmed2021,causalworldcode} & A known-PD mechanism in physical units, integrated with learned mass-affine terms and an object-state predictor.\\
Evaluation & Controlled simulation and input ablation & Disjoint link-mass ranges, separately trained mass-information controls, and comparison of ID and OOD mass benefits.\\
\bottomrule
\end{tabularx}
\end{table}

\section{Complementary Benchmark Details}
\label{app:benchmark}
\subsection{Environment-transfer predictors and metric}
The environment-transfer data record simulator state variables and physical descriptors together with nine action coordinates. Constant source features are removed, and remaining inputs and output observations are standardized using source-training statistics. The two tasks use ten seeds each. Per run, 20\% of the 8,000 source examples form an ID test set; the remaining examples are split 85:15 for model fitting and validation. A separate pool of 4,000 paired source interventions estimates dependencies. The effect threshold is 0.03 times the largest measured effect, and action dependencies are kept open because the paired pool intervenes on state variables rather than actions. No target prediction scores are used in model validation.

The PETS-style baseline has five bootstrapped probabilistic MLPs, each with two hidden layers of width 200 and SiLU activations. Each network predicts a mean and diagonal log variance, is trained with Gaussian negative log likelihood, and uses Adam with learning rate $10^{-3}$, batch size 512, at most 300 epochs, and validation patience 30. Point predictions average the five network means. Its graph variant retains an input if the estimated mask connects it to at least one output; the masked predictor instead uses output-specific dependencies. All variants use the same source observations and target observations. Graph variants additionally use the independent source intervention pool, and the comparison evaluates their complete predictive configurations.

The intervention-masked predictor uses an eight-dimensional scalar encoding, 16-dimensional edge messages, and encoder and output-head hidden widths of 64. It is trained with squared prediction error, Adam at learning rate $3\times10^{-3}$, batch size 512, at most 300 epochs, and validation patience 30. The reported 0.6122 configuration does not add a paired-response loss. The paired data are used to estimate its mask, and the mask is held fixed during predictor training.

For standardized target observation $\tilde y_n$ and prediction $\hat{\tilde y}_n$, the reported score is
\begin{equation}
\mathcal S=1-\frac{\sum_n\|\hat{\tilde y}_n-\tilde y_n\|_2^2}
{\sum_n\|\tilde y_n-\bar{\tilde y}_{\mathrm{train}}\|_2^2},
\end{equation}
where $\bar{\tilde y}_{\mathrm{train}}$ is the source-training mean. This measures prediction skill against the source-training constant predictor; Eq.~\eqref{eq:metric} uses the per-step test mean for joint $R^2$. Task means and the overall mean aggregate run-level scores. The three benchmark rows share this definition and target set.

\subsection{RWM implementation and shared-window evaluation}
\label{app:rwm}
The comparison adapts the official RWM state-dynamics module~\cite{li2025rwm,rwmcode} at commit \nolinkurl{18eebcdd7145284c8d5eed5d8ed1a4b96c649693}. Its two-layer, 256-unit GRU feeds separate mean and log-standard-deviation heads, each with a 128-unit ReLU hidden layer. The adapter uses the official sampled autoregressive regression and variance-bound losses. The input has 59 coordinates: 31 states, nine actions, 12 robot masses, and seven object/environment context values. No-mass variants zero all 12 standardized mass slots during training and evaluation, retaining the input dimension and initialization of their actual-mass counterparts. This benchmark evaluates the state predictor; policy optimization and auxiliary reward or termination heads are outside its prediction task.

RWM receives 32 observed state--action pairs and trains to forecast either 8 or 50 future states. Adam uses learning rate $10^{-4}$, weight decay $10^{-5}$, batch size 1,024, a gradient-norm limit of 1.0, and variance-bound weight 1.0. Each configuration has a budget of 50,000 gradient updates, with ID validation every 250 updates. Early stopping uses 20 validation checks without improvement after at least 2,000 updates. Selection minimizes joint NRMSE over the common 100-step ID validation forecast. Both training horizons use this selection protocol. The four structured configurations use their existing validation-selected checkpoints from Table~\ref{tab:training}.

For evaluation, RWM is initialized with $(s_1,a_1),\ldots,(s_{32},a_{32})$, producing $\hat s_{33}$; later steps feed predicted states and the recorded next action back into the model. Structured models start directly from $s_{32}$. All models predict $s_{33}$ through $s_{132}$ without future observed states. RWM feeds back its predicted mean at evaluation, and the shared evaluation applies quaternion normalization. The hold-state reference repeats $s_{32}$ at every forecast step. All 300 OOD episodes and 240 ID episodes are evaluated per task--seed run. Non-finite predictions count as failures. Table~\ref{tab:rwmappendix} separates the task medians and reports ID accuracy under the same window.

\begin{table}[!htbp]
\centering\small
\caption{Additional shared-window scores. ID columns aggregate ten task--seed runs; each OOD task column aggregates its five seeds. All scores are median 100-step mean joint $R^2$ (higher is better).}
\label{tab:rwmappendix}
\begin{tabularx}{\linewidth}{Xlccc}
\toprule
Model & Mass input & ID & OOD pushing & OOD picking\\
\midrule
Hold-state reference & None & -0.332 & -0.327 & -0.316\\
\midrule
RWM (8) & None & 0.642 & 0.715 & 0.719\\
RWM (8) & Actual & 0.626 & -0.850 & -0.447\\
\midrule
RWM (50) & None & 0.635 & 0.752 & 0.751\\
RWM (50) & Actual & 0.622 & -0.658 & -0.421\\
\midrule
Mass-affine, learned torque (A) & Mean & 0.701 & 0.764 & 0.774\\
Mass-affine, learned torque (A) & Actual & 0.693 & 0.751 & 0.756\\
\midrule
Mass-affine, known PD (B) & Mean & 0.620 & 0.661 & 0.682\\
Mass-affine, known PD (B) & Actual & 0.635 & 0.745 & 0.754\\
\bottomrule
\end{tabularx}
\end{table}

\end{document}